%% file: main_revised.tex
\documentclass[conference]{IEEEtran}
\IEEEoverridecommandlockouts
\usepackage[english]{babel}
\usepackage{cite}
\usepackage{amsmath,amssymb,amsfonts}
\usepackage{algorithmic}
\usepackage{acronym}
\usepackage{booktabs}
\usepackage{graphicx}
\usepackage{textcomp}
\usepackage{xcolor}
\usepackage{tikz}
\usepackage{multirow}
\usepackage{lipsum}
\usepackage{makecell}

\usetikzlibrary{positioning}
\def\BibTeX{{\rm B\kern-.05em{\sc i\kern-.025em b}\kern-.08em
    T\kern-.1667em\lower.7ex\hbox{E}\kern-.125emX}}

\input{ACRONYMS.tex}

\begin{document}

\title{Combining Convolution and Delay Learning in Recurrent Spiking Neural Networks}


\author{\IEEEauthorblockN{Lúcio Folly Sanches Zebendo}
\IEEEauthorblockA{
lucio.follysancheszebendo@studenti.unipd.it}
\and
\IEEEauthorblockN{Eleonora Cicciarella}
\IEEEauthorblockA{
eleonora.cicciarella@phd.unipd.it}
\and
\IEEEauthorblockN{Michele Rossi$^\dag$}
\IEEEauthorblockA{
michele.rossi@unipd.it} 
\thanks{All authors are with the University of Padova, \textit{Department of Information Engineering}. $^\dag$This author is also with the \textit{Department of Mathematics} ``Tullio-Levi Civita'' at the University of Padova, Italy. 
}}

\maketitle

\begin{abstract}
Spiking neural networks (SNNs) are rapidly gaining momentum as an alternative to conventional artificial neural networks in resource constrained edge systems. In this work, we continue a recent research line on recurrent SNNs where axonal delays are learned at runtime along with the other network parameters~\cite{queant2025delrec}. The first proposed approach, dubbed DelRec, demonstrated the benefit of recurrent delay learning in SNNs. Here, we extend it by advocating the use of convolutional recurrent connections in conjunction with the DelRec delay learning mechanism. According to our tests on an audio classification task, this leads to a streamlined architecture with smaller memory footprint (around $99\%$ savings in terms of number of recurrent parameters) and a much faster ($52\times$) inference time, while retaining DelRec's accuracy. Our code is available at: https://github.com/luciozebendo/delrec\_snn/tree/conv\_delays
\end{abstract}


\section{Introduction}

Sequential data play a central role in many real-world applications, including speech recognition, gesture analysis, and biomedical signal processing, such as ECG and EEG recordings. These data are characterized by complex temporal dependencies across multiple time scales, which require models that can integrate information over time while preserving fine-grained temporal structure. 

\acp{snn} provide a biologically inspired framework for temporal processing by encoding information in the form of sparse, asynchronous binary events, known as \emph{spikes}. Their event-driven computation naturally aligns with the structure of temporal data streams, enabling high responsiveness  and reduced computational overhead. 
To model long-range temporal dependencies, \acp{snn} are commonly extended with recurrence, forming \acp{rsnn}. Through recurrent connections and neuron state dynamics, \acp{rsnn} maintain an internal memory that allows them to integrate information over extended time horizons and exhibit rich temporal behavior. Nevertheless, training \acp{rsnn} on complex temporal tasks using gradient-based optimization remains challenging. Although the use of surrogate gradients~\cite{neftci2019surrogate} enables backpropagation through time, gradients can still vanish or explode over long time sequences, as in standard recurrent networks.

Recent research has primarily focused on improving \acp{rsnn} performance by enhancing spiking neuron models, for instance through adaptive temporal dynamics~\cite{bittar2022surrogate, baronig2025advancing}. 
In contrast, a promising alternative is the incorporation of \emph{learnable delays}, a mechanism that is already observed in biological systems. 
In this context, a delay represents the time required for a spike emitted by a presynaptic neuron to reach a postsynaptic neuron. Delays enhance the network’s computational power by enabling neurons to respond selectively to specific relative spike timings. Indeed, when incoming connections carry different delays, neurons can represent complex temporal sequences that would be difficult to capture with uniform or zero delays. From a training perspective, delays can also be interpreted as skip connections in time, facilitating the flow of gradients and improving the network’s ability to learn long-range temporal dependencies.

The benefits of delay learning have been extensively demonstrated in feedforward \acp{snn}~\cite{deckers2024colearning, goltz2025delgrad}, and more recent studies have begun to explore their use in recurrent architectures. For instance, the authors of~\cite{xu2025asrcsnn} investigated learning a single recurrent delay per layer using backpropagation, demonstrating the effectiveness of parameterized delay mechanisms for long-term temporal modeling.
Similarly, Queant \emph{et al.}~\cite{queant2025delrec} proposed \emph{DelRec}, a \ac{rsnn} incorporating learnable synaptic delays to reactivate past signals, by enhancing temporal modeling without increasing neuron complexity and achieving state-of-the-art performance on the \ac{ssc} dataset.

Building on DelRec, in this paper we extend axonal delay learning to \acp{crsnn}. Indeed, DelRec uses \emph{fully dense} recurrent layers, but the strong local correlations present in many temporal signals make dense connectivity unnecessary and potentially inefficient. By introducing convolutional recurrence, our approach exploits this locality to reduce parameter overhead while maintaining expressive temporal modeling. 

\noindent The main contributions of this work are summarized as follows:
\begin{itemize}
    \item We introduce a convolutional recurrent architecture for DelRec~\cite{queant2025delrec}, replacing \emph{fully dense} recurrent connections with lightweight 1D convolutions. This modification leverages local correlations in temporal signals, drastically reducing parameter overhead.
    \item We show that local connectivity, combined with learnable axonal delays, is sufficient for effective temporal modeling in \acp{rsnn}, achieving state-of-the-art accuracy on the \ac{shd} audio dataset, while reducing recurrent parameters by $99\%$ and accelerating inference by more than $50\times$ compared to the original DelRec implementation.
    \item Through an ablation study, we show that learnable delays improve accuracy and training stability compared to fixed delays, highlighting their critical role in capturing long-range temporal dependencies.

\end{itemize}




\section{Background}\label{sec:backgroung}

\subsection{Spiking Neural Networks}

\acp{snn} are biologically inspired neural networks where neurons communicate using discrete binary signals, referred to as \textit{spikes}, rather than continuous, real-valued activations used in standard \acp{ann}. This event-driven computing property makes \acp{snn} especially suitable for neuromorphic hardware, where energy is drained only when spikes are fired~\cite{eshraghian2023training}, enabling significant energy savings compared to \acp{ann}~\cite{neuromorphic2011zhou}.

The most widely used model of spiking neuron is the \ac{lif} neuron, as it balances biological realism and computational efficiency. At a given discrete time step $t$, the neuron maintains a membrane potential $V[t]$, which integrates an input current $I[t]$ while exponentially decaying toward a resting value. The membrane potential \textit{before reset} (or the hidden state) is denoted as $H[t]$ and defined as:
\begin{equation}
    H[t] = \beta \cdot V[t-1] + (1 - \beta) \cdot I[t],
    \label{eq:lif_charge}
\end{equation}
where $\beta = 1 - 1/\tau$ is the membrane decay factor and $\tau$ is the membrane time constant. This hidden state $H[t]$ represents the accumulated charge at time $t$ before the spiking logic is applied. 
A spike $S[t]$ is emitted when $H[t]$ exceeds a predefined threshold $V_{\text{th}}$, according to:
\begin{equation}
    S[t] = \Theta(H[t] - V_{\text{th}}),
    \label{eq:spike}
\end{equation}
where $\Theta(\cdot)$ is the Heaviside step function. 
After firing, the membrane potential is reset. Two typical reset schemes are  the \textit{hard reset}, which sets $V[t] = H[t] (1-S[t])$, meaning that when a spike occurs ($S[t]=1$) the membrane potential is reset to zero, and the \textit{soft reset}, which subtracts the threshold from the hidden state, i.e., $V[t] = H[t] - V_{\text{th}} S[t]$, allowing the neuron to retain residual charge after firing.

Training \acp{snn} with gradient-based methods is challenging because the Heaviside function in Eq.~\eqref{eq:spike} has zero derivative almost everywhere, preventing direct backpropagation. \ac{sgl}~\cite{neftci2019surrogate} overcomes this issue by replacing, during the backward pass, the true derivative of the Heaviside function with a smooth approximation $\Theta^\prime(x) \approx f^\prime(x)$, where $f(\cdot)$ is a differentiable surrogate function, such as the arctangent or triangular functions. This enables the use of standard backpropagation through time to optimize \ac{snn} parameters, making it the predominant supervised training approach for modern \acp{snn}~\cite{eshraghian2023training}.

\subsection{Learnable delays in Recurrent Spiking Neural Networks}

In biological neural networks, signals require time to propagate between neurons: axonal conduction introduces temporal delays that vary across connections, influenced by factors such as myelination and axon length. These propagation latencies are functional, allowing neurons to detect temporal coincidences among incoming spikes and facilitating phenomena like polychronization (i.e., the ability to produce precisely timed, non-synchronous sequences of firing events that can arise from strongly connected groups of neurons)~\cite{queant2025delrec}.

Incorporating learnable delays into \ac{snn} connections in computational models has been shown to greatly improve their temporal processing capabilities. In a feedforward setting, the \ac{dcls} method~\cite{hammouamri2024dcls} showed that optimizing synaptic delays improves performance on temporal benchmarks. Intuitively, \textit{recurrent} delays are more impactful: they allow for self-sustained neuronal activity, enable long-range temporal dependencies to be formed and exploited, and can potentially {\it mitigate vanishing gradients} by creating temporal skip connections in the computational graph.

DelRec~\cite{queant2025delrec} proposes a novel idea, a \ac{sgl}-based method for learning axonal delays in recurrent connections. In a standard \ac{rsnn}, the recurrent input to neuron $i$ at time $t$ is computed as
\begin{equation}
    X_i^{\text{rec}}[t] = \sum_{j} W_{ij}^{\text{rec}} \cdot S_j[t - 1],
    \label{eq:rsnn_vanilla}
\end{equation}
where $W^{\text{rec}}$ is the matrix storing the recurrent weights and all connections have a fixed unit delay. DelRec extends this by assigning each neuron $j$ a learnable axonal delay $d_j \in \mathbb{N}$, so that spikes reach connected neurons at time $t + 1 + d_j$ (rather than at $t + 1$):
\begin{equation}
    X_i^{\text{rec}}[t] = \sum_{j} W_{ij}^{\text{rec}} \cdot S_j[t - (1 + d_j)].
    \label{eq:rsnn_delayed}
\end{equation}
To optimize the model, DelRec considers real-valued delays and proposes a differentiable triangular spread function $h_{\sigma,d}(\tau)$ that spreads each spike's influence over adjacent time steps:
\begin{equation}
    h_{\sigma,d}(\tau) = \max\!\left(0,\; \frac{1 + \sigma - |\tau - (1 + d)|}{(1 + \sigma)^2}\right).
    \label{eq:spread}
\end{equation}
During training, the width parameter $\sigma$ is annealed from a large initial value down to zero, progressively sharpening the spread from a broad distribution to a linear interpolation between the two nearest integers. Finally, at inference time the delays are rounded to the nearest integer.

From an implementation perspective, this is realized using a circular buffer (called the \textit{scheduling matrix}) of shape $N \times L_{\text{buf}}$, where $N$ is the number of neurons and $L_{\text{buf}}$ depends on the maximum delay and current $\sigma$. At each time step, spikes are weighted and scheduled into future buffer positions according to the spread function, and the current buffer entry provides the recurrent input.

Using this approach with simple \ac{lif} neurons, DelRec achieved state-of-the-art results on the Spiking Speech Commands (SSC) dataset and competitive performance on the \ac{shd} dataset~\cite{cramer2022heidelberg}, demonstrating that learnable recurrent delays are a powerful mechanism for temporal processing in \acp{snn}. A key observation from this work is that the recurrent weight matrix $\mathbf{W}^{\text{rec}} \in \mathbb{R}^{N \times N}$ introduces $N^2$ parameters per layer, which becomes prohibitive as layers grow wider. Our proposed method addresses this scalability concern, as detailed in the following section.

\vspace{4pt}
\section{Proposed Method: Convolutional Recurrence}\label{sec:method}


The {\it dense} recurrent weight matrix $\mathbf{W}^{\text{rec}} \in \mathbb{R}^{N \times N}$ in the original DelRec architecture generates a quadratic ($N^2$) number of parameters per layer. For example, for a layer with $N = 256$ neurons, this results in $65{,}536$ parameters per layer exclusively for the recurrent information, a major overhead that scales quadratically with the layer size.

We observe that signals often do contain correlation in space-time and in these cases dense layers may be a suboptimal choice. For example, the audio spectrograms used in datasets like \ac{shd} and \ac{ssc} display sound as time-frequency representations where each input channel corresponds to a frequency band from a cochlear model. Adjacent frequency channels usually exhibit correlated patterns, due to the harmonic structure of speech signals, the spectral continuity of acoustic events, and the overlapping filter responses in cochlear models.
This local structure suggests that {\it full all-to-all connectivity may be redundant}. The rationale is that each neuron, representing a frequency channel, mainly needs to interact with its immediate neighbors (local connectivity) rather than with all other neurons in the layer. This often better reflects the correlation underpinning the data.

\subsection{Convolutional recurrent connections}

In this work, we advocate replacing the {\it dense} recurrent matrix with a lightweight one-dimensional (1D) convolution kernel:
\begin{equation}
    \mathbf{W}^{\text{rec}} \in \mathbb{R}^{N \times N} \quad \longrightarrow \quad \mathbf{W}^{\text{conv}} \in \mathbb{R}^{k},
\end{equation}
with $k$ denoting the kernel size. For instance, we use $k = 3$ throughout this work, meaning that each neuron receives recurrent input only from itself and its two immediate neighbors.
The modified spike scheduling equation becomes:
\begin{equation}
    \mathbf{X}^{\text{rec}}[t+\tau] = \mathbf{W}^{\text{conv}} * \left(\mathbf{h}_{\sigma,\mathbf{d}}(\tau) \odot \mathbf{S}[t]\right),
    \label{eq:conv_schedule}
\end{equation}
where $\odot$ represents the element-wise multiplication and $*$ denotes a 1D convolution along the neuron frequency dimension. It is worth noting that, following the convention of modern deep learning frameworks such as PyTorch, the $*$ operator specifically implements \textit{cross-correlation}. Unlike the formal mathematical definition of convolution, the kernel is not flipped before the sliding window operation, which is the standard practice in convolutional neural network implementations.

The delay learning mechanism remains unchanged: each neuron has its own learnable delay $d_j$, and the triangular spread function is used to enable differentiable optimization. The core modification lies in the spatial connectivity pattern, which shifts from a global to a local structure.

\subsection{Implementation details}

To efficiently implement the 1D convolutional recurrent kernel alongside the existing buffer mechanism, we cast it as a 2D convolution with kernel shape  $(k, 1)$. Let $B$ denote the batch size, $N$ the number of neurons, and $L$ the buffer length. The input tensor of shape $(B, N, L)$ is first reshaped to  $(B, 1, N, L)$ by adding a channel dimension. Then, the convolutional kernel only convolves the input along the $N$ dimension. A stride of 1 and zero-padding are applied to preserve dimensionality. Finally, the output is reshaped back to $(B, N, L)$ by removing the extra dimension.
The convolutional kernel is initialized using Kaiming uniform distribution, while the delays follow the original half-normal or uniform initialization from DelRec. For clarity, we still refer to this operation as a 1D convolution throughout the paper.

\subsection{Parameter analysis}

Table~\ref{tab:param_comparison} compares the parameter counts between the original ({\it dense}) architecture from DelRec and our modified version that employs convolutions.

\vspace{-8pt}
\begin{table}[htbp]
\caption{Number of recurrent parameters per layer.}
\begin{center}
\begin{tabular}{lcc}
\toprule
\textbf{Component} & \textbf{DelRec (Dense)} & \textbf{Ours (Conv.)} \\
\midrule
Recurrent weights & $N^2$ & $k$ \\
Axonal delays & $N$ & $N$ \\
\midrule
\textbf{Total} & $N^2 + N$ & $k + N$ \\
\bottomrule
\end{tabular}
\label{tab:param_comparison}
\end{center}
\end{table}
\vspace{-8pt}
\noindent For $N = 256$ neurons and a kernel size $k = 3$, DelRec requires  $256^2 + 256 = 65{,}792$ recurrent parameters, whereas our convolutional approach requires only $3 + 256 = 259$, corresponding to a remarkably $\mathbf{99.6\%}$ \textbf{reduction in the total number of parameters}.
The axonal delays ($N$ parameters per layer) are preserved, as they encode the \textbf{temporal} scheduling mechanism, which is crucial to capture long-range dependencies in the time domain. The parameter reduction targets only the \textbf{spatial} connectivity, which we argue is largely redundant for locally-structured inputs.

\vspace{4pt}
\section{Network Architecture}\label{sec:network}

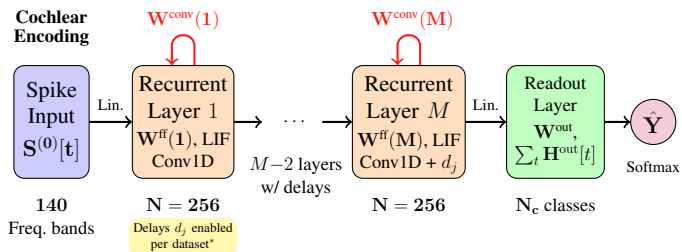
\begin{figure}[tbp]
    \centering
    \begin{tikzpicture}[scale=0.5, every node/.style={scale=0.5}]
      
      \node[draw, rectangle, fill=blue!20, minimum width=2.cm, minimum height=3cm, rounded corners] (input) at (-5, 0) {};
      \node[align=center, font=\LARGE] at (input.center) {Spike\\Input\\[5pt]$\mathbf{S^{(0)}[t]}$};
      \node[below=0.2cm, font=\Large, align=center] at (input.south) {$\mathbf{140}$ \\[0.5pt] {Freq.  bands}};
      \node[above=0.2cm, font=\Large, align=center] at (input.north) {\textbf{Cochlear}\\[-0.1cm]\textbf{Encoding}};
      
      \node[draw, rectangle, fill=orange!25, minimum width=2.7cm, minimum height=3cm, rounded corners] (layer1) at (-1.5, 0) {};
      \node[align=center, font=\LARGE] at ([yshift=0.6cm]layer1.center) {Recurrent\\Layer $1$};
      \node[font=\Large, align=center] at ([yshift=-0.7cm]layer1.center) {$\mathbf{W^{\text{ff}}(1)}$, LIF\\Conv1D};
      \node[below=0.2cm, font=\Large] at (layer1.south) {{$\mathbf{N=256}$}};
      
      \node[fill=yellow!40, rounded corners, font=\normalsize, align=center, minimum width=2.2cm] at (-1.5, -3.05) {Delays $d_j$ enabled\\per dataset$^*$};
      
      \draw[->, thick] (input.east) -- node[above, yshift=5pt, font=\large] {Lin.} (layer1.west);
      
      \node[font=\Large] at (1.5, 0) {$\cdots$};
      \node[below=0.01cm, font=\Large, align=center] at (1.5, -0.8) {$M{-}2$ layers\\w/ delays};
      
      \draw[->, thick] (layer1.east) -- (0.6, 0);
      
      \node[draw, rectangle, fill=orange!25, minimum width=3cm, minimum height=3cm, rounded corners] (layerL) at (4.5, 0) {};
      \node[align=center, font=\LARGE] at ([yshift=0.6cm]layerL.center) {Recurrent\\Layer $M$};
      \node[font=\Large, align=center] at ([yshift=-0.8cm]layerL.center) {$\mathbf{W^{\text{ff}}(M)}$, LIF\\Conv1D + $d_j$};
      \node[below=0.2cm, font=\Large] at (layerL.south) {{$\mathbf{N=256}$}};
      
      \draw[->, thick] (2.2, 0) -- (layerL.west);
      
      \node[draw, rectangle, fill=green!25, minimum width=2.6cm, minimum height=3cm, rounded corners] (readout) at (8.4, 0) {};
      \node[align=center, font=\Large] at ([yshift=0.7cm]readout.center) {Readout\\Layer};
      \node[font=\Large, align=center] at ([yshift=-0.6cm]readout.center) {$\mathbf{W}^{\text{out}}$,\\ $\sum_t \mathbf{H}^{\rm out}[t]$};
      \node[below=0.2cm, font=\Large] at (readout.south) {$\mathbf{N_c}$ classes};
      
      \draw[->, thick] (layerL.east) -- node[above, yshift=5pt,, font=\large] {Lin.} (readout.west);
      
      \node[draw, circle, fill=purple!20, minimum size=1.2cm] (output) at (11, 0) {};
      \node[font=\LARGE] at (output.center) {$\hat{\mathbf{Y}}$};
      \node[below=0.15cm, font=\large] at (output.south) {Softmax};
      
      \draw[->, thick] (readout.east) -- (output.west);
      
      \draw[->, thick, red!90, rounded corners] 
        ([xshift=0.3cm]layer1.north) -- ++(0, 0.8) -| ([xshift=-0.3cm]layer1.north);
      \node[font=\Large, red!90] at (-1.5, 2.7) {$\mathbf{W^{\text{conv}}(1)}$};
      
      \draw[->, thick, red!90, rounded corners]
        ([xshift=0.3cm]layerL.north) -- ++(0, 0.8) -| ([xshift=-0.3cm]layerL.north);
      \node[font=\Large, red!90] at (4.6, 2.7) {$\mathbf{W^{\text{conv}}(M)}$};
      
      
    \end{tikzpicture}
    \caption{Overview of the complete network architecture. The input spike trains ($140$ frequency bands after cochlear binning) are processed by $M$ hidden recurrent layers with \ac{lif} neurons. Each layer employs 1D convolutional recurrence ($\mathbf{W}^{\rm conv}$, kernel size $k=3$) with learnable axonal delays $d_j$. The delay configuration of the first layer is dataset-specific. A non-spiking readout layer integrates the membrane potentials $\mathbf{H}[t]$ over time, and the predicted class $\hat{\mathbf{Y}}$ is obtained through a softmax operation.}

    \label{fig:full_architecture}
\end{figure}

The network follows a feedforward-recurrent architecture featuring multiple hidden layers. A detailed diagram is depicted in  Figure~\ref{fig:full_architecture}. Each hidden layer consists of a linear projection (with weights $\mathbf{W}^{\rm ff}$), followed by batch normalization, a layer of \ac{lif} neurons, a convolutional recurrent delay unit and, finally, a dropout regularization.
The output layer consists of a linear classifier followed by a layer of \ac{lif} neurons with an infinite threshold, which acts as a leaky integrator for readout purposes. In this layer, no spikes are generated; instead, the prediction of the network is obtained by applying a softmax function to membrane potentials at the last time step.

\subsection{Convolutional recurrent delay unit}
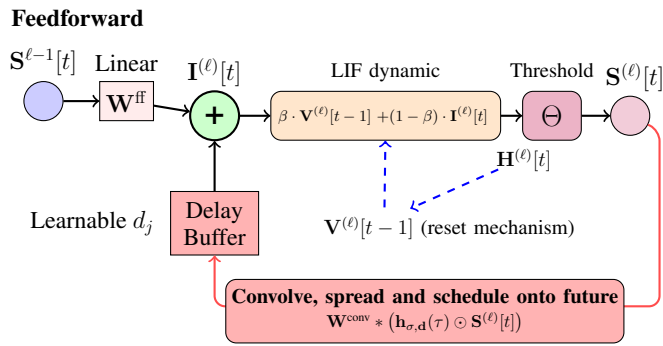
\begin{figure}[tbp]
    \centering
    \begin{tikzpicture}[scale=0.65, every node/.style={scale=0.65}]
      \node[font=\Large] at (-4.8, 2.2) {\textbf{Feedforward}};
      \node[font=\Large] at (-5.5, 1.3) {$\mathbf{S}^{\ell-1}[t]$};
      \node[draw, circle, fill=blue!20, minimum size=0.8cm] (input_ff) at (-5.5, 0.5) {};
      
      \node[draw, rectangle, fill=pink!30, minimum width=1.1cm, minimum height=0.8cm] (Wff) at (-3.8, 0.5) {};
      \node[font=\Large] at (Wff.center) {$\mathbf{W}^{\rm ff}$};
      \node[above=0.05cm, font=\Large] at (Wff.north) {Linear};
      \draw[->, thick] (input_ff) -- (Wff);
      
      \node[draw, rectangle, fill=red!30, minimum width=1.8cm, minimum height=1.3cm] (delay_buf) at (-2, -2) {};
      \node[align=center, font=\Large] at (delay_buf.center) {Delay\\[-0.05cm]Buffer};
      \node[font=\Large] at (-4.5,-2) {Learnable $d_j$};
      
      
      \node[draw, circle, minimum size=1cm, fill=green!20, line width=0.8pt] (sum) at (-2, 0.2) {};
      \node at (sum.center) {\huge\textbf{+}};
      \draw[->, thick] (Wff) -- (sum);
      \draw[->, thick] (delay_buf) -- (sum);
      
      \node[right=0.3cm, above=0.25cm, font=\Large] at (-2,0.3) {$\mathbf{I}^{(\ell)}[t]$};
      
      
      \node[draw, rectangle, rounded corners, fill=orange!20, minimum width=4.7cm, minimum height=1.0cm] (membrane) at (1.5, 0.2) {};
      \node[align=center, font=\normalsize] at (membrane.center) {
        \small $\beta \cdot \mathbf{V}^{(\ell)}[t-1]$ \small $+(1-\beta) \cdot \mathbf{I}^{(\ell)}[t]$
      };
       \node[above=0.05cm, font=\large] at (membrane.north) {LIF dynamic};
      \node[align=center, font=\large] at (4.4,-0.7) {
        $\mathbf{H}^{(\ell)}[t]$ 
      };
      \draw[->, thick] (sum) -- (membrane);
      
      
      \draw[->, dashed, thick, blue] (1.47, -2) -- (membrane);
      \node[fill=white, font=\large] at (2.8, -2.1) {$\mathbf{V}^{(\ell)}[t-1]$ (reset mechanism)};
      \draw[->, dashed, thick, blue] (3.8,-0.9) -- (2,-1.7);
      
      \node[draw, rectangle, rounded corners, fill=purple!30, minimum width=1.2cm, minimum height=1.cm] (spike_gen) at (4.9, 0.2) {};
      \node[align=center, font=\LARGE] at (spike_gen.center) {$\Theta$\\[-0.7cm]};
      \node[above=0.1cm, font=\large] at (spike_gen.north) {Threshold};
      \draw[->, thick] (membrane) -- (spike_gen);
      
      \node[draw, circle, fill=purple!20, minimum size=0.8cm] (output) at (6.5, 0.2) {};
      \draw[->, thick] (spike_gen) -- (output);
      \node[font=\Large] at (6.6, 1) {$\mathbf{S}^{(\ell)}[t]$};
      
      
      \draw[->,  thick, red!70, rounded corners] 
        (output) -- ++(0.6, -0.3) 
        |- ++(-0.5, -3.6) 
        -| (delay_buf);
      
      \node[draw, rectangle, fill=red!30, rounded corners, align=center, minimum width=2.2cm, minimum height=1.3cm] at (2.3, -3.8) {
        \large \textbf{Convolve, spread and schedule onto future}\\[0.07cm]
        $\mathbf{W}^{\text{conv}} * \left(\mathbf{h}_{\sigma,\mathbf{d}}(\tau) \odot \mathbf{S}^{(\ell)}[t]\right)$
      };
      
      
    \end{tikzpicture}
    \caption{Design of the convolutional recurrent delay unit. The feedforward path consists of a simple linear projection. The recurrent path replaces the dense weight matrix of \cite{queant2025delrec} with 1D convolution in the neuron dimension with kernel size $k=3$, while keeping the learnable delay buffer mechanism.}
    \label{fig:architecture}
\end{figure}

Figure~\ref{fig:architecture} illustrates the recurrent block of the proposed architecture at the generic layer $\ell$. We denote by $\mathbf{S}^{(\ell)}$ the output spike signal of layer $\ell$ at time $t$.

At each time step, the layer receives two contributions. First, the feedforward input from the previous layer, $\ell-1$, undergoes a linear transformation parametrized by weights $\mathbf{W}^{\text{ff}}$.
Second, recurrent processing is performed by retrieving the previously emitted spikes from the circular delay buffer. The output spike signal is then spread through the function $h_{\sigma,d_j}(\tau)$, convolved along the neuron dimension with the convolutional kernel $\mathbf{W}^{\rm conv} \in \mathbb{R}^{k}$, and subsequently scheduled into future buffer positions. 
The total input current $\mathbf{I}^{(\ell)}[t]$, obtained by summing the feedforward and recurrent contributions, drives the \ac{lif} dynamics according to Eq.~\eqref{eq:lif_charge}.

\vspace{5pt}
\section{Experimental Validation}\label{sec:results}

\subsection{Datasets}

The proposed architecture and delay learning mechanism are evaluated on two neuromorphic audio benchmarks from the Heidelberg Spiking Datasets~\cite{cramer2022heidelberg}:
\begin{itemize}
    \item The Spiking Heidelberg Digits (SHD) dataset, which contains $10{,}420$ recordings of spoken digits ($0-9$) in English and German, resulting in $20$ classes. The official split consists of $8{,}156$ training samples and $2{,}264$ test samples. To enable hyperparameter tuning, we randomly partition the original training set into an $80\%$ training subset and a $20\%$ validation subset, while keeping the original test set unchanged.

    \item The Spiking Speech Commands (SSC) dataset, which comprises $105,829$ recordings of $35$ different spoken words. It is provided with a predefined training, validation, and test split of $75{,}466$ training samples, $9{,}981$ validation samples, and $20{,}382$ test samples.
\end{itemize}

\noindent In both datasets, the audio signals are encoded into spike trains using a $700$-band cochlear model, which we aggregate into $140$ frequency bins, following the original DelRec preprocessing.

\subsection{Training configuration}

For both datasets, network training is performed using \ac{sgl} with the arctangent function. The loss function is the standard cross-entropy, computed between the output of the readout layer and the true label.
Network hyperparameters for the \ac{shd} dataset are optimized using Ray Tune with Optuna, while for the SSC dataset, we adopt the baseline parameters from~\cite{queant2025delrec} without additional tuning. Other training-related hyperparameters, including parameters specific to the \ac{lif} neurons, are summarized in Table~\ref{tab:config}.

\vspace{-7pt}
\begin{table}[htbp]
\caption{Training configurations.}
\vspace{-8pt}
\begin{center}
\begin{tabular}{lcc}
\toprule
\textbf{Parameter} & \textbf{SHD} & \textbf{SSC} \\
\midrule
Optimizer & AdamW & Adam \\
Learning rate weights / delays & 0.0013 / 0.0279 & 0.001 / 0.05 \\
Dropout ff. / rec. & 0.44 / 0.26  & 0.1 / 0.3  \\
Membrane time constant $\tau$ & 1.17 & 2.0  \\
$\sigma_{\text{init}}$ / $\sigma_{\text{decay}}$  & 10.36 /  0.971 & 10.0 / 0.95\\
Reset mechanism & Hard& Soft \\
SNN time steps & 100 & 250 \\
\bottomrule
\end{tabular}
\label{tab:config}
\end{center}
\end{table}
\vspace{-15pt}
\subsection{Numerical results}

We trained and tested our convolutional recurrent delay learning architecture on the \ac{shd} dataset with 2 and 4 hidden layers to evaluate network scalability. To enable a direct comparison with DelRec\cite{queant2025delrec}, we also trained a corresponding 4-layer DelRec model, since the original paper only implements a 2-layer architecture. For the \ac{ssc} dataset, we kept the number of layers identical to the original DelRec setup, i.e., 3 layers.

As shown in Table~\ref{tab:results}, on the SHD dataset, our convolutional approach achieves an accuracy of $91.51$\% $\pm$ $0.70$\%, closely matching the results in the original DelRec paper ($91.72$\% $\pm$ $0.84$\%). The accuracy gap of just $0.21$ percentage points is remarkably small considering that our convolutional variant reduces the recurrent parameters by $\mathbf{99.995}$\% (with the chosen configurations, from $65,536$ to $3$ parameters per layer). Moreover, the lower standard deviation ($0.70$\% {\it vs} $0.84$\%) suggests more stable training dynamics with convolutional recurrence. These results indicate that local connectivity captures nearly all the temporal information encoded by dense recurrent connections, while being orders of magnitude more parameter-efficient. 
We also show that the original DelRec does not scale well with the number of layers, as accuracy decreases when moving from 2 to 4 layers, likely due to overfitting caused by the increased number of parameters.
Remarkably, on top of the smaller memory footprint, our method {\it substantially reduces the inference time} ($52$ times faster), resulting extremely suitable for online streaming applications.

On SSC, our method achieves $78.59$\% $\pm$ $0.39$\% using baseline hyperparameters from the original paper, showing a gap of almost 4 percentage points compared to the original DelRec results.

The learned delay statistics are reported in Table~\ref{tab:stats}. For SHD, we compare our convolutional model with four hidden layers against the original DelRec architecture with two layers. Overall, both approaches exhibit similar mean delay values across datasets and layers, indicating that they learn comparable central temporal scales despite the difference in depth. However, differences emerge in the dispersion of the learned delays. In particular, our model tends to produce broader distributions, as reflected by consistently higher standard deviations and wider ranges, especially on SHD. This suggests a richer and more heterogeneous allocation of delays across neurons. In contrast, DelRec shows more concentrated delay distributions with lower variability around the mean.
\vspace{-8pt}
\begin{table}[htbp]
\caption{Performance comparison on SHD and SSC test sets.}
\vspace{-9pt}
\begin{center}
\scalebox{0.93}{
\begin{tabular}{llcccc}
\toprule
\textbf{Dataset} & \textbf{Method} & \textbf{Layers} &
\textbf{\makecell{Test Acc. \\ {[\%]} }} &
\textbf{\makecell{Rec.\\Params.}} &
\textbf{\makecell{Inference\\Time {[ms]}}}
 \\
\midrule
\multirow{2}{*}{SHD} 
  & DelRec* & 2 & 91.72 $\pm$ 0.84 & $65,536$ & 37.11 \\
  & Ours & 2 & 89.32 $\pm$ 1.29 & $3$ & 0.71 \\
\midrule
\multirow{2}{*}{SHD} 
  & DelRec & 4 & 90.41 $\pm$ 0.83 & $196,608$ & 38.03 \\
  & Ours & 4 & \textbf{91.51 $\pm$ 0.70} & $9$ & 1.51  \\
\midrule
\multirow{2}{*}{SSC} 
  & DelRec$\dag$ & 3 & 82.58 $\pm$ 0.08 & $196,608$ & 112.64 \\
  & Ours & 3 & 78.59 $\pm$ 0.39 & $9$ & 4.19 \\
\bottomrule

\multicolumn{6}{l}{\footnotesize
\parbox[t]{\linewidth}{
$^*$ Results reproduced using publicly available code.\\
$\dag$ Results reported in the original paper.\\
Accuracies are reported as mean $\pm$ standard deviation over $9$ (SHD) and $3$ (SSC) different seeds.
}}
\end{tabular}
}
\label{tab:results}
\end{center}
\end{table}
\vspace{-15pt}

\vspace{-7pt}
\begin{table}[htbp]
\caption{Delay statistics across layers for our model and DelRec.}
\vspace{-8pt}
\begin{center}
\scalebox{0.95}{
\begin{tabular}{llcccccc}
\toprule
\textbf{} & \textbf{} & \multicolumn{3}{c}{\textbf{Ours}} & \multicolumn{3}{c}{\textbf{DelRec}} \\
\cmidrule(lr){3-5} \cmidrule(lr){6-8}
\textbf{Dataset} & \textbf{Layer} & \textbf{Mean} & \textbf{Std} & \textbf{Range} & \textbf{Mean} & \textbf{Std} & \textbf{Range} \\
\midrule
\multirow{3}{*}{SHD} 
& $\ell = 1$ & 21.13 & 12.17 & [0, 62] & 20.78 &  7.67 & [5, 41] \\
& $\ell = 2$ & 21.90 & 12.98 & [0, 72] & -- & -- & -- \\
& $\ell = 3$ & 18.74 & 10.24 & [0, 43] & --  & -- & -- \\
\midrule
\multirow{3}{*}{SSC} 
& $\ell = 1$ & 10.41 & 8.87 & [0, 51] & 11.34  & 8.29 & [0, 47] \\
& $\ell = 2$ & 7.51  & 9.88 & [0, 60] & 8.62  & 9.93 & [0, 50] \\
& $\ell = 3$ & 9.75 & 9.53 & [0, 57] & 9.74 & 9.23 & [0, 44] \\
\bottomrule
\end{tabular}
}
\label{tab:stats}
\end{center}
\end{table}
\vspace{-15pt}

\subsection{Ablation study on the importance of learnable delays}

To assess the effectiveness of learning delays, we perform an ablation study comparing our full model with two fixed-delay (e.g., non-learnable) variants. In the first variant, all delays are set to $d_j = 1$ and kept constant during training, reducing the model to a standard recurrent architecture. In the second variant, delays are fixed to either the mean or median of the learned delay distribution, in order to evaluate whether a representative statistic of the learned delays can preserve performance. In both cases, all other model components remain unchanged. 

The results on both datasets are reported in Table~\ref{tab:ablation_delays}. In general, we observe a consistent performance drop when replacing learnable delays with fixed ones. The most pronounced case is on SHD, where removing delay learning leads to a decrease of more than $5$ percentage points in accuracy, highlighting the importance of adaptive temporal dynamics.


\begin{table}[h!]
\caption{Effect of fixed vs. learnable delay configuration.}
\begin{center}
\begin{tabular}{llcc}
\toprule
\textbf{Dataset} & \textbf{Configuration} & \textbf{Test Accuracy [\%]} & \textbf{Seeds} \\
\midrule
\multirow{3}{*}{SHD} 
 & Fixed ($d_j = 1$) & $86.28 \pm 4.82$ & 10 \\
 & Fixed ($d_j = 21$)$^{\dagger\ddagger}$ & $88.19 \pm 0.65$ & 10 \\
 & Learnable & $\mathbf{91.51 \pm 0.70}$ & 10 \\
\midrule
\multirow{3}{*}{SSC} 
& Fixed ($d_j = 1$) & $75.09 \pm 2.82$ & 3 \\
& Fixed ($d_j = 7$)$^\ddagger$ & $77.65 \pm 0.37$ & 3 \\
& Learnable & $\mathbf{78.59 \pm 0.39}$ & 3 \\
\bottomrule
\multicolumn{4}{l}{\footnotesize $^\dagger$Mean of learned delay distribution.} 
\\
\multicolumn{4}{l}{\footnotesize $^\ddagger$Median of learned delay distribution.}
\end{tabular}
\label{tab:ablation_delays}
\end{center}
\end{table}
\vspace{-15pt}



\section{Conclusions}\label{sec:conclusion}




In this paper, we propose a convolutional recurrent \ac{snn} variant of DelRec that preserves learnable axonal delays while replacing \emph{dense} recurrent connections with lightweight 1D convolutions ($k=3$). This leverages local correlations in audio spectrograms, where adjacent channels exhibit correlated activations patterns due to speech harmonics.

Our results show that local connectivity combined with learnable delays is sufficient for effective temporal modeling in \acp{rsnn}. On the \ac{shd} dataset, our model matches DelRec accuracy while reducing recurrent parameters by $>\mathbf{99\%}$ and achieving $\mathbf{52\times}$ faster inference time. An ablation study confirms the importance of delay learning, improving accuracy by more than $5$ percentage points over fixed delays on \ac{shd}.

These findings suggest that the dense recurrent connectivity in DelRec introduces substantial redundancy for signals with local spatial structure. Our convolutional formulation scales to wider layers without quadratic cost, making it suitable for deployment on resource-constrained neuromorphic hardware.

\bibliography{references}
\bibliographystyle{ieeetr}
\end{document}

%% file: acronyms.tex
\acrodef{ann}[ANN]{Artificial Neural Network}
\acrodef{crsnn}[CRSNN]{Convolutional Recurrent Spiking Neural Network}
\acrodef{dcls}[DCLS]{Dilated Convolution with Learnable Spacings}
\acrodef{lif}[LIF]{Leaky Integrate-and-Fire}
\acrodef{rsnn}[RSNN]{Recurrent Spiking Neural Network}
\acrodef{ssc}[SSC]{Spiking Speech Commands}
\acrodef{shd}[SHD]{Spiking Heidelberg Digits}
\acrodef{snn}[SNN]{Spiking Neural Network}
\acrodef{sgl}[SGL]{Surrogate Gradient Learning}